\documentclass{llncs}
\usepackage{graphicx}
\usepackage{pgfplots}
\usepackage{tikz}
\usepackage{pstricks,pst-plot,pst-text,pst-tree,pst-eps,pst-fill,pst-node,pst-math}
\usetikzlibrary{shapes}
\usetikzlibrary{calc}
\usepackage{tabularx}
\usepackage{latexsym}
\usepackage{algorithm,algorithmic}
\usepackage{url}
\makeatletter
\newenvironment{figurehere}
{\def\@captype{figure}}
{}
\makeatletter

\usepackage{makeidx}  
\begin{document}
\title{Climbing Depth-bounded Adjacent Discrepancy Search for Solving
  Hybrid Flow Shop Scheduling Problems with Multiprocessor Tasks}
\author{Asma LAHIMER \inst{1} \and Pierre LOPEZ \inst{1} \and Mohamed
  HAOUARI \inst{2} }
\institute{
CNRS ; LAAS ; 7 avenue du colonel Roche, F-31077 Toulouse Cedex 4, France \\
Universit\'e de Toulouse ; UPS, INSA, INP, ISAE ; UT1, UTM, LAAS ; F-31077 Toulouse Cedex 4, France \\
\email{asma.lahimer@laas.fr, pierre.lopez@laas.fr}
\and
INSAT ; Institut National des Sciences Appliqu\'ees et de
Technologie \\ Centre Urbain Nord BP 676 - 1080 Tunis Cedex, Tunisie \\
\email{mohamed.haouari@insat.rnu.tn}
}
\maketitle

\begin{abstract}
  This paper considers multiprocessor task scheduling in a multistage
  hybrid flow-shop environment. The problem even in its simplest form
  is NP-hard in the strong sense. The great deal of interest for this
  problem, besides its theoretical complexity, is animated by needs of
  various manufacturing and computing systems. We propose a new
  approach based on limited discrepancy search to solve the
  problem. Our method is tested with reference to a proposed lower
  bound as well as the best-known solutions in
  literature. Computational results show that the developed approach
  is efficient in particular for large-size problems.

  \keywords{Hybrid flow shop scheduling, Multiprocessor tasks, Discrepancy search}
\end{abstract}

\section{Introduction}
\label{intro}

Flow shop scheduling refers to a manufacturing facility in which all
jobs visit the production machines in the same order. In hybrid flow
shop scheduling, the jobs serially traverse stages following the same
production route, and must be assigned to one of the parallel machines
composing each stage.  The hybrid flow shop scheduling problem with
multiprocessor tasks is itself a generalization of the hybrid flow
shop problem, allowing tasks to be processed on more than one
processor in a given stage, at a time. It can also be viewed as a
specific case of the resource-constrained project scheduling problem
(RCPSP).

Many applications of hybrid scheduling problems with multiprocessor
tasks can be found in various manufacturing systems (\emph{e.g.},
work-force assignment in \cite{Chen1999}, transportation problem with
recirculation in \cite{Bertel2004}), as well as in some computer
systems (\emph{e.g.}, real-time machine-vision \cite{Ercan1999}).

Hybrid flow shop scheduling problem with multiprocessor tasks has
received considerable attention from researchers and has been solved
by various approaches, \emph{e.g.} genetic algorithms \cite{oguz05},
tabu search, and ant colony system \cite{serifoglu06}. Motivated by
the success of discrepancy search for solving shop scheduling
problems, in particular hybrid flow shop \cite{benhmida10},
\cite{benhmida07}, we propose in this paper a new approach based on
discrepancy search to solve the hybrid flow shop problem with
multiprocessor tasks.


\section{Problem Definition}
\label{sec:1}

The hybrid flow shop scheduling problem with multiprocessor tasks can
be formally described as follows: A set \textbf{J}=\{$1,2,\ldots,n$\}
of $n$ jobs, have to be processed in $m$ stages. Hence, a job is a
sequence of $m$ tasks (one task for each stage). Each stage
$\textbf{i}=\{1,2,\ldots,m\}$ consists of $m_{i}$ identical parallel
processors. In a stage $i$, the job $j$ requires simultaneously
$size_{ij}$ processors. That is, $size_{ij}$ processors selected at
stage $j$ are required for processing job $j$ for a period of time
equal to the processing time requirement of job $j$ at stage $i$,
namely $p_{ij}$. The objective is to minimize the \emph{makespan}
($C_{\max} $), that is, the completion time of all tasks in the last
stage. According to the classical 3-field notation in production
scheduling, the problem is denoted by
Fm({$m_{1}$},$\ldots$,{$m_{m}$})$\vert${$size_{ij}$}$\vert${$C_{\max}$}.


\section{Discrepancy Search}
\label{sec:2}

\subsection{General Statement}

Limited discrepancy search (LDS) was introduced in $1995$ by Harvey
and Ginsberg \cite{harvey95}. This seminal method can be considered as
an alternative to the branch-and-bound procedure, backtracking
techniques, and iterative sampling.  From an optimization view-point
this technique is similar to variable neighbourhood search. Indeed, it
starts from an initial global instantiation suggested by a given
heuristic and successively explores branches with increasing
discrepancies from it, in order to obtain a solution (in a
satisfaction context), or a solution of better performance (in an
optimization context).  A discrepancy is associated with any decision
point in a search tree where the choice goes against the heuristic.
For convenience, in a tree-like representation the heuristic choices
are associated with left branches while right branches are considered
as discrepancies.  Since LDS proposition in 1995, several variants
were suggested, among them, Improved Limited Discrepancy Search (ILDS)
\cite{kor96}, Depth-bounded Discrepancy Search (DDS) \cite{walsh97},
Discrepancy-Bounded Depth First Search \cite{Beck00} and Climbing
Discrepancy Search (CDS) \cite{milano02}. \\

In the following sections, we focus on those methods that inspired our
approach, in particular DDS and CDS.


\subsection{Depth-bounded Discrepancy Search}

Depth-bounded Discrepancy Search (DDS) developed in \cite{walsh97}, is
an improved LDS that prioritizes discrepancies at the top of the tree
to correct early mistakes first.  This assumption is ensured by means
of an iteratively increasing bound on the tree depth. Discrepancies
below this bound are prohibited. DDS starts from an initial
solution. At $ith$ iteration, it explores those solutions on which
discrepancies occur at a depth not greater than $i$.

\subsection{Climbing Discrepancy Search}

Climbing Discrepancy Search (CDS) is a local search method adapted to
combinatorial optimization problems proposed in \cite{milano02}. CDS
starts from an initial solution that would be dynamically
updated. Indeed, it visits branches progressively until a better
solution is reached. Then, the initial solution is updated and the
exploration process is restarted.


\section{Proposal: Climbing Depth-bounded Adjacent Discrepancy Search}

\subsection{CDADS: Main Features}

To stick to the problem under consideration, we now consider an
optimization context. We propose CDADS (Climbing Depth-bounded
Adjacent Discrepancy Search) method, that is a combination of a
depth-bounded discrepancy search and a climbing discrepancy search. We
also assume that, if several discrepancies occur in the construction
of a solution, these discrepancies are necessarily \emph{adjacent} in
the list of successive decisions. CDADS starts from an initial
solution obtained by a given heuristic, and explores its neighborhood
progressively, according to the depth-bounded discrepancy search
strategy. Hence, a limit depth $d$ is fixed. Discrepancies below this
bound are prohibited. At the $ith$ iteration, we allow $i$
discrepancies above the limit level $d$. \\

When considering solutions with more than one discrepancy, we require
these discrepancies are achieved consecutively, that means a solution
consists of discrepancies that happen one after the other. This
assumption of adjacency considerably limits the search space. We also
consider that the initial solution is generated by a `good'
heuristic. Thus, only the immediate neighborhood of a discrepancy may
receive an additional discrepancy. We obtain a truncated DDS based on
adjacent discrepancies, DADS (Depth-bounded Adjacent Discrepancy
Search).  This approach is illustrated by an example on a binary tree
of depth 3 (see Figure~\ref{figure1}).  At the starting point, DADS
visits the initial solution recommended by the heuristic. For
convenience, we assume that left branches follow the heuristic. At
first iteration, DADS visits leaf nodes at the depth limit with
exactly one discrepancy. The first line shown under the branches
reports the visit order of considered solution, while the second line
illustrates the number of discrepancies made in each solution. The 2nd
iteration allows to exploring more solutions with two discrepancies
with respect to the adjacency assumption. In this representation, the
maximum depth bound is taken to be $3$. If now, we limit the depth to
two levels, several branches would not be retained, namely the
branches $4$, $6$, and $7$ would not be visited by DADS.

\begin{center}
\begin{figurehere}
\begin{tikzpicture}[level 1/.style={level distance= 1.2cm,sibling distance=3.5cm},
                    level 2/.style={level distance= 1.5cm,sibling distance=1.8cm},
                    level 3/.style={level distance= 1.1cm,sibling distance=0.9cm},
                    scale=0.5
                     ]
\begin{scope}[every node/.style={draw,ellipse,inner sep=1pt,fill=black}]

\node (root) {}
  
   child {[very thick]node {}
         child {node {}
                child {node {}}
                child {[thin]node {}}
               }
         child {[thin]node {}
                child {[thin]node {}}
                child {[thin]node {}}
               }
        }
  child {node {}
         child {[thin]node {}
                child {node {}}
                child {[thin]node {}}
               }
         child {node {}
                child {node {}}
                child {[thin]node {}}
               }
        }
  
  ;
 
\end{scope}

   \path let \p{root} =(root)
   in    
       node [scale=0.8]at (\x{root},\y{root}+0.7cm) {\textit{0th Iteration}}  
       node [left,scale=0.7] at (\x{root}-2.9cm,\y{root}-4.4cm) {1}
       node [left,scale=0.7] at (\x{root}-2.9cm,\y{root}-4.8cm) {0};

  \begin{scope}[every node/.style={draw,ellipse,inner sep=1pt,fill=black},shift={(9,0)}]

\node (root) {}
  
   child {[very thick]node {}
         child {node {}
                child {[thin]node {}}
                child {node {}}
               }
         child {node {}
                child {node {}}
                child {[thin]node {}}
               }
        }
  child {[very thick]node {}
         child {node {}
                child {node {}}
                child {[thin]node {}}
               }
         child {[thin]node {}
                child {node {}}
                child {[thin]node {}}
               }
        }
  
  ;
 
\end{scope}
\path let \p{root} =(root)
   in    
       node [scale=0.8]at (\x{root},\y{root}+0.7cm) {\textit{1st Iteration}}

      node [left,scale=0.7] at (\x{root}-1.9cm,\y{root}-4.8cm) {1}
      node [left,scale=0.7] at (\x{root}-1.9cm,\y{root}-4.4cm) {4}
      
      node [left,scale=0.7] at (\x{root}-1.1cm,\y{root}-4.8cm) {1}
      node [left,scale=0.7] at (\x{root}-1.1cm,\y{root}-4.4cm) {3}
      
       node [left,scale=0.7] at (\x{root}+0.7cm,\y{root}-4.8cm) {1}
      node [left,scale=0.7] at (\x{root}+0.7cm,\y{root}-4.4cm) {2}
      
      ;

      \begin{scope}[every node/.style={draw,ellipse,inner sep=1pt,fill=black},shift={(0,-8)}]

\node (root) {}
  
   child {[very thick]node {}
         child {[thin]node {}
                child {node {}}
                child {node {}}
               }
         child {node {}
                child {[thin]node {}}
                child {node {}}
               }
        }
  child {[very thick]node {}
         child {[thin]node {}
                child {node {}}
                child {node {}}
               }
         child {[very thick]node {}
                child {node {}}
                child {[thin]node {}}
               }
        }
  
  ;

\end{scope}
\path let \p{root} =(root)
   in    
       node[scale=0.8] at (\x{root},\y{root}+0.7cm) {\textit{2nd Iteration}}

      node [left,scale=0.7] at (\x{root}-0.1cm,\y{root}-4.8cm) {2}
      node [left,scale=0.7] at (\x{root}-0.1cm,\y{root}-4.4cm) {6}

      node [left,scale=0.7] at (\x{root}+2.5cm,\y{root}-4.8cm) {2}
      node [left,scale=0.7] at (\x{root}+2.5cm,\y{root}-4.4cm) {5}

      ;

 \begin{scope}[every node/.style={draw,ellipse,inner sep=1pt,fill=black},shift={(9,-8)}]

\node (root) {}
  
    child {node {}
         child {node {}
                child {node {}}
                child {node {}}
               }
         child {node {}
                child {node {}}
                child {node {}}
               }
        }
  child {[very thick]node {}
         child {[thin]node {}
                child {node {}}
                child {node {}}
               }
         child {[very thick]node {}
                child {[thin]node {}}
                child {node {}}
               }
        }
  
  ;
 
\end{scope}
\path let \p{root} =(root)
   in    
      node [scale=0.8] at (\x{root},\y{root}+0.7cm) {\textit{3th Iteration}}  
     
      node [left,scale=0.7] at (\x{root}+3.5cm,\y{root}-4.8cm) {3}
      node [left,scale=0.7] at (\x{root}+3.5cm,\y{root}-4.4cm) {7}

      ;
     
\end{tikzpicture}

\caption{Depth-bounded Ajacent Discrepancy Search}

\label{figure1}
\end{figurehere}
\end{center}

Going back to the optimization issue, CDADS merges the DADS strategy
with a CDS exploration principle, that is the initial solution used by
DADS is dynamically updated when a best solution is found, and the
exploration process is restarted.

\subsection{Heuristics}
\label{heuristics}

CDADS is strongly based on the quality of the initial solution. Thus,
we carried out an experimental comparison between various priority
rules presented in the literature \cite{serifoglu06}, \cite{oguz03}.
We considered the most effective heuristics to multiprocessor task
hybrid flow shop scheduling.
The four selected rules are:
\begin{itemize}
\item \textbf{SPT} (Shortest Processing Time), which ranks jobs
  according to the ascending order of their processing times;
\item \textbf{SPR} (Shortest Processing Requirement), which ranks jobs
  according to the ascending order of their processing requirement;
\item the \textbf{Energy} rule, considering first the jobs with the
  smallest energy (where the energy of an operation $j$ at a stage $i$
  is evaluated by $p_{ij} \times size_{ij}$); and
\item \textbf{NSPT\_LastStage} (Normalized SPT applied at the last
  stage). For this latest rule, \c{S}erifo\u{g}lu and Ulusoy
  \cite{serifoglu06} propose to schedule jobs according to their
  ranking index ($RI_{j}$) defined by:
  \begin{center}
    $RI_{j} = \frac{\displaystyle \max_k \{p_{mk}\} -
      p_{mj}+1}{\displaystyle \max_k \{p_{mk}\} + 1}$.
  \end{center}
\end{itemize}

In Table~\ref{selectionH}, the selected priority rules are ranked
according to their percentage of best solutions found, that is,
performance.

\begin{table}[ht]
  \centering
  \caption{Heuristic selection}
  \label{selectionH}

  \begin{tabular}{cc}
    \tabularnewline
    \hline 
    Priority Rule & Performance (\%) \\
    \hline \\[-3mm]
    NSPT\_LastStage & 27 \\[1mm]
    Energy& 25 \\[1mm]
    SPT & 17 \\[1mm]
    SPR & 14 \\
    \hline

  \end{tabular}
\end{table}


\subsection{Schedule Generation Scheme}

Schedule generation schemes (SGSs) are widely used in solving
preemptive problems. We distinguish between serial SGS and parallel
SGS. These two heuristics ensure task scheduling based on a given
priority rule. Hence, tasks are selected one after the other and a
start time is fixed for each one.

Serial SGSs are introduced in \cite{kel63}.  At each iteration, the
first available task in $\zeta$ is selected, where $\zeta$ is the
priority list recommended by the priority rule. The selected task is
scheduled as soon as possible with respect to both resource
constraints and precedence constraints.

Parallel SGSs developed in \cite{bro65}, suggest a chronological
procedure in scheduling tasks. At each time $t$, a set $\zeta_{t}$ of
tasks being scheduled is defined: this set contains unscheduled tasks
that can be processed at $t$ without breaking neither precedence
constraints nor resource constraints. If we consider that
$\underline{t}$ is the first time where $\zeta_{\underline{t}}\neq
\emptyset$, the first task in the priority list $\zeta$ belonging to
$\zeta_{\underline{t}}$ is performed at $\underline{t}$. The same
process is applied until all tasks are scheduled. The two schemes
depicted above may appear similar. However, the schedule they generate
are different: a serial SGS provides an active schedule while a
parallel SGS generates a non-delay schedule.

In the scheduling theory, Sprecher \emph{et al.} \cite{spr95} show
that the set of active schedules includes at least one optimal
solution. On the contrary, non-delay schedules may eliminate all
optima.

Concerning our method CDADS, we do not enumerate all possible
solutions, so even serial SGSs may exclude all optimum
solutions. Furthermore, in practice, parallel SGSs are known for their
operational efficiency. Hence, we opt for the implementation of a
parallel SGS which has been proved, moreover, to be more efficient in
our experimental studies.

\subsection{Lower Bound}

For efficiency purpose, we join CDADS with an evaluation of lower
bounds at each node. The proposed lower bound is based on lower bounds
previously presented in \cite{oguz05}. Thus, we suggest this formula:
\begin{center}
  $LB=\max(LB^{s}$,$ LB^{j})$
\end{center}
where $LB^{j}$ is a job-based lower bound similar to the one suggested in
\cite{oguz05}: $LB^{j}=\displaystyle \max_{j \in J}(\sum_{i=1}^{m}p_{ij})$; 
and $LB^{s}$ is a stage-based lower bound: $LB^{s}$=$\displaystyle
\max_{i=1..m}LB(i)$. \\

\noindent
For this latter bound, we claim that:
\[ LB(i)= \left \{
  \begin{array}{ll}
    \max[M_{1}(i),M_{2}(i),\displaystyle \max_{j\in J}(p_{ij})] +
    \min_{j\in J}(\sum_{l=i+1}^{m}p_{lj}) \; , & \forall i=1 \\
    \displaystyle\min_{j\in J}(\sum_{l=1}^{i-1}p_{lj}) +
    \max[M_{1}(i),M_{2}(i),\displaystyle \max_{j\in J}(p_{ij})] +
    \min_{j\in J}(\sum_{l=i+1}^{m}p_{lj}) \; , & \forall i=2..m-1 \\
    \displaystyle\min_{j\in J}(\sum_{l=1}^{i-1}p_{lj}) +
    \max[M_{1}(i),M_{2}(i),\displaystyle \max_{j\in J}(p_{ij})] \; , & \forall i=m
  \end{array}
\right . \]
where
\[ M_{1}(i)=\biggl \lceil \displaystyle \frac{1}{m_{i}}\sum_{j \in
  J}(p_{ij}size_{ij}) \biggr \rceil \]
\centerline{and}
  \[ M_{2}(i)=\displaystyle \sum_{j\in A_{i}}p_{ij}+\frac{1}{2}\sum_{j\in B_{i}}p_{ij} \; , \]
with
\[ A_{i}=\displaystyle \lbrace j \vert size_{ij} > \frac{m_{i}}{2}\rbrace \]
\centerline{and}
\[ B_{i}=\displaystyle \lbrace j \vert size_{ij} = \frac{m_{i}}{2}\rbrace. \]


\noindent
\emph{Justification of the expression of $LB(i)$}.
\begin{quote}
  We assume that only non-delay task scheduling is considered. \\[-2mm]

  The first term of $LB(i)$ gives a lower bound on the beginning of
  every job $j \in J$ on any machine of stage $i$. \\[-2mm]

  The last term can be explained accordingly, since it is associated
  with the minimal required time to achieve the processing of every
  job $j$ on all the subsequent stages of stage $i$. \\[-2mm]

  The middle term concerns the processing of jobs on stage $i$.
  $M_1(i)$ stands for the mean stage load for job preemptive
  scheduling, while $M_2(i)$ reviews two different situations for
  partitionning the jobs according to their resource requirement. Set
  $A_i$ consists of jobs that must be processed sequentially (resource
  requirement greater than the half of the resource capacity
  $m_i$). Set $B_i$ groups together the jobs having a resource
  requirement exactly equal to the half of the resource
  capacity. Obviously, a job belonging to $A_i$ and another job
  belonging to $B_i$ must also be processed sequentially.  The added
  term $\displaystyle \max_{j\in J}(p_{ij})$ contributes to maximize
  the evaluation of stage load on a considered stage $i$, especially
  when some jobs having high processing time are being
  scheduled. \\[-1mm]

  This justifies the validity of the bound.  \hfill $\Box$
\end{quote}

\section{Computational Study}
\label{experiments}

\subsection{Test Beds}

For comparison purpose, we assess the performance of CDADS on
instances of O\u{g}uz's benchmark available on her home page:
\url{http://home.ku.edu.tr/coguz/public_html/}. This benchmark is
widely used in the literature \cite{sivrikaya04}, \cite{jouglet09},
\cite{oguz04}. \\

The number of jobs is taken to be $n=5,10,20,50,100$ and the number of
stages $m$ takes its value from the set \{$2,5,8$\}.  The benchmark
considers two types of problems, ``Type-1'' and ``Type-2''. In
`Type-1' instances, the number of processors $m_{i}$ available at each
stage $i$ (resource capacity) is randomly determined from the set
\{$1,\ldots,5$\}, while in `Type-2' $m_{i}$ is fixed to $5$ processors
for every stage $i$. In fact, `Type-2' instances are globally more
flexible than `Type-1 instances'. For each combination of $n$ and $m$,
and for each type, $10$ instances are randomly generated, which leads
a total of $300$ instances. The processing time of each job $j$ in
stage $i$ ($p_{ij}$) and its processing requirement ($size_{ij}$) are
integer and are randomly generated from sets \{$1,\ldots,100$\} and
\{$1,\ldots,m_{i}$\}, respectively. \\

The algorithm implementing CDADS was coded in C++ and run on an Intel
core 2 Duo 2 GHz PC. The maximum CPU time is set to 60 seconds. The
exploration is also stopped when CDADS reaches a given lower bound on
the makespan. Obviously, if CDADS misses the optimal solution, the
best-found solution when the maximum CPU time is reached, is then
taken to be the problem solution.


\subsection{Restart Policy}
\label{restart}

For the computational study, we have then retained four priority rules
to generate the initial solutions (see Section~\ref{heuristics}). That
is why whe have introduced a restart policy to benefit from these
heuristics.  At a starting point, we use the best rule, that is the
NSPT\_LastStage. However, if no improvement is noticed during the
CDADS search, we restart the process with another solution obtained by
applying the next rule ``Energy'' that could lead a more efficient
solution for this specific instance, and so on.

The restart policy is limited by the size of the heuristics pool:
restarts are then allowed at most four times, since we have selected
four rules. At each restart $k$ (starting from $k=0$), we increase the
number of maximum nodes that can be visited according to a geometrical
series \emph{nbrNodes}$\,\times f^{k}$, where $f$ is fixed to $1.3$
and \emph{nbrNodes} varies linearly with the problem size (the number
of jobs $n$; for example for $n=20$ we fix \emph{nbrNodes} to $2000$
nodes). Hence the search space is expanded at each restart.


\subsection{Results}

We tested two strategies for applying discrepancy: Top First and
Bottom First.  In the Top First exploration, discrepancies at the top
of the tree are privileged while the Bottom First strategy favors
discrepancies at the bottom. Computational study shows that CDADS is
really more efficient with a Top First strategy (then contradicting
--\,for the problem at hand\,-- the statement of relative indifference
of discrepancy order by~\cite{Prosser08}). Thus, the results shown
below refer to this latter strategy. \\

Table~\ref{PerformanceCDADS} gives for each configuration ($n$: number
of jobs, and $m$: number of stages) and each type, the average
percentage deviation (\textit{\%dev}) and the average CPU time.  The
average percentage deviation is measured in two ways:
\begin{itemize}
\item[$\bullet$] For small problems, solutions are compared to the
  optimal solutions ($C_{\max}^*$ denotes the optimum makespan):
  \[\displaystyle \frac{C_{\max}-C_{\max}^{*}}{C_{\max}^{*}} \times
  100;\]
\item[$\bullet$] For larger problems, solutions found by the CDADS are
  compared to the lower bound ($LB$):
  \[\displaystyle \frac{C_{\max}-LB}{LB} \times 100.\]
\end{itemize}

As explained in Section~\ref{restart}, CDADS is run four times on each
of the selected priority rules (NSPT\_LastStage, Energy, SPT, SPR) for
each instance. The best solution is taken to be the CDADS solution for
the corresponding problem. According to findings of
\cite{serifoglu06}, the
Fm({$m_{1}$},$\ldots$,{$m_{m}$})$\vert${$size_{ij}$}$\vert${$C_{\max}$}
problem and its symmetric have the same optimal makespan. Referring to
this property, we apply a two-directional planning (forward schedule
and backward schedule).


\begin{table}[ht]
  \centering
  \caption{CDADS performance}
  \label{PerformanceCDADS}
  
  \begin{tabular}{@{}ccccp{3mm}cc}
    \hline 
    \multicolumn{2}{l}{} & \multicolumn{2}{l}{`Type-1' Problems} &&
    \multicolumn{2}{l}{`Type-2' Problems} \\
    \cline{3-4}  \cline{6-7}
    $n$ & $m$&  \begin{small}\textit{\%dev}\end{small} & \begin{small}\textit{CPU(s)} \end{small}  && \begin{small}\textit{\%dev}\end{small} & \begin{small}\textit{CPU(s)} \end{small} \\
    \cline{1-7} \\[-2mm]
    $5$ & $2$ &  0 & $<0.1$ &&  0 & $<0.1$ \\
    & $5$ &  0.21 & $<0.1$   && 0.46 & $<0.1$ \\
    & $8$ &  1.71& $<0.1$  && 0.5 & $<0.1$ \\
	
    \tabularnewline
    $10$ & $2$ &  0 & $<0.1$ &&   1.72 & $<0.1$ \\
    & $5$ &  0.66 & 0.4   && 6.44 & $<0.1$ \\
    & $8$ &  8.47 & $<0.1$ &&   9.61 & 0.2 \\
	
    \tabularnewline
    $20$ & $2$ &  0.05 & 0.1 &&   3.34 & 3.1 \\
    & $5$ &  2.57 & 1.1 &&   7.97&1.3\\
    & $8$ &  5.11 & 0.2 &&   15&1.3 \\
	
    \tabularnewline
    $50$ & $2$ & 0.49 & 2.3 &&   1.74 & 4.2 \\
    & $5$ &  0.54& 5 &&   8.2 &13.5\\
    & $8$ &  1.62 & 6.8   && 12.42 &33.4 \\
	
    \tabularnewline
    $100$ & $2$ &  0.08 & 11.1  && 3.32 & 22.8 \\
    & $5$ &  1.5 & 13.6  && 10.75 &40.9 \\
    & $8$ &  1.86 & 11 && 14.33 &47.3 \\
    \cline{1-7}
    \multicolumn{2}{l}{\textit{Global average}}& 1.66 & 3.44  && 6.39 & 10.53 \\
    \cline{1-7}

  \end{tabular}
\end{table}

From Table~\ref{PerformanceCDADS}, it is observed that the average
percentage deviation is higher for `Type-1' instances. Globally,
\textit{\%dev} is 1.66\% for `Type-1' problems and 6.39\% for `Type-2'
problems. This increase can be linked to several assumptions: the
lower bound becomes less effective as $m_{i}$ increases in `Type-2'
instances and so the average percentage deviation would be
higher. Another explanation can also be considered: the number of
processors are fixed in `Type-2' problems, that is $m_{i}=5$, and the
scheduling problem becomes more difficult to solve for CDADS.

Results show the behavior of our approach with variations of $n$ and
$m$.  For a given $n$, the average percentage deviation increases with
increasing $m$. Indeed, the problem difficulty increases when $m$
increases and the obtained solution is further away from the lower
bound. On the other hand, for a given number of stages $m$, increasing
$n$ has no significant effect on the average percentage deviation, as
the effectiveness of CDADS is independent of the number of jobs: the
stability of our method seems to be not linked to the number of jobs
$n$, since for a given $m$ (\emph{e.g.}, $m=8$), in `Type-1' problems,
when $n$ increases from $50$ jobs to $100$ jobs, the average
percentage deviation increase slightly (from $1.62\%$ to $1.86\%$).
It also can be noticed, that in some cases, increasing $n$ results in
a decrease in the deviation value (for the configuration $n=20, m=8$
the \textit{\%dev} is taken to be $5.11\%$, and is evaluated to
$1.62\%$ for $n=50, m=8$). Apparently, the lower bound becomes more
effective with $n$ increasing.

From the experimental studies, it can be observed that CDADS converges
quickly. The average CPU time varies between less than $0.1$ seconds
and $47.3$ seconds. The computational cost is more important in
`Type-2' instances, confirming the difficulty of these
problems. Similarly, for a fixed $m$, increasing $n$ leads to CPU time
increase. Conversely, when $n$ is fixed, increasing $m$ increases the
CPU time.

\subsection{Comparison of CDADS Solutions with State-Of-the-Art Results}

Table~\ref{Comparaison} presents the results of CDADS on
\textit{\%dev}, the average percentage deviation (as well as a
synthesis of the average CPU time for all instances, in the last line
of the table). Furthermore, it shows the results obtained by Jouglet
\emph{et al.} in \cite{jouglet09}. These results are the most recent
and the best-known solutions in literature. Thus, we have compared the
results of CDADS with GA (genetic algorithm), CP (constraint
programming), and MA (a memetic algorithm that combines GA and CP). We
disregard the results published by Ercan \emph{et al.}  \cite{oguz05}
given inconsistency encountered. We contrast our results only versus
those presented in \cite{jouglet09}. However, we omit the average
deviation published in this latest paper due to detected
miscalculation (induced by Ercan \emph{et al.}'s errors). Hence, we
recalculated the average percentage deviation for all methods given in
\cite{jouglet09}. The maximum CPU time is fixed at $900$ seconds for
GA, CP, and MA.

\begin{table}[ht]
  \centering
  \caption{Comparing average percentage deviation (and CPU time)}
  \label{Comparaison}

  \begin{tabular}{c c c c c c c c c c c c c}
    \hline 
    \multicolumn{3}{c}{} & \multicolumn{4}{c}{`Type-1' Problems} & & &
    \multicolumn{4}{c}{`Type-2' Problems} \\
    \cline{4-7}  \cline{10-13}
    $n$ & $m$& & \begin{small}\textit{CDADS}\end{small} & \begin{small}\textit{GA} \end{small} & \begin{small}\textit{CP}\end{small} & \begin{small}\textit{MA} \end{small}& & &\begin{small}\textit{CDADS}\end{small} & \begin{small}\textit{GA} \end{small} & \begin{small}\textit{CP}\end{small} & \begin{small}\textit{MA} \end{small}\\
    \cline{1-13} \\[-2mm]
    $5$ & $2$ & & 0 & 0.29 & 0 & 0  	& & &	0 & 1.23 & 0 & 0 \\
    & $5$ & & 0.21 &1.35 &0&0	 	& & &	0.46 & 1.44 & 0 & 0\\
    & $8$ & & 1.71& 4.15 &	0&0 	& & &	0.5&2.38& 0&0\\
	
    \tabularnewline
    $10$ & $2$ & & 0 & 0&0&0 			& & & 	1.72 & 2.83&1.72&1.75 \\
    & $5$ & & 0.66 &1.64& 0 &0 		& & &	6.44 &7.8&6.1&5.67 \\
    & $8$ & & 8.47&9.38 &10.32&8.02 & & & 	9.61 &10.87&8.37&8.8 \\
	
    \tabularnewline
    $20$ & $2$ & & 0.05 & 0.44&2.59&0.66 & & &	3.34 & 3.7&6.72&3.43 \\
    & $5$ & & 2.57 & 3.49&10.85&2.78 & & &  7.97&9.57&22.86&9.57\\
    & $8$ & & 5.11 & 5.69&17.98&5.32 & & &  15&17.26&28.52&16.02 \\
	
    \tabularnewline
    $50$ & $2$ & & 0.49 & 0.63&2.79&0.49 & & & 1.74 & 2.76&6.54&2.21 \\
    & $5$ & & 0.54 & 0.59 &5.3&0.51  & & & 8.2 &10.95&20.01&10.32\\
    & $8$ & & 1.62 & 2.17 &14.42&1.71& & & 12.42 &15.89&30.06&17.25 \\
	
    \tabularnewline
    $100$ & $2$ & & 0.08 & 0.15 &1.96&0.07 & & & 3.32 & 3.05&5.68&2.7 \\
    & $5$ & & 1.5 & 2.5 & 5.19 &2.33 & & &10.75&14.95&19.13&14.37 \\
    & $8$ & & 1.86 & 1.99&9.47&2.15  & & & 14.33 &20.06&23.15&17.83 \\
    \cline{1-13}
    \multicolumn{2}{c}{\textit{Global average}}& & \textbf{1.66} & 2.27 &	5.39& 1.6& & & \textbf{6.39} &7.28&11.92&8.32  \\
    \cline{1-13} \\[-2mm]
    \cline{1-13}
    \multicolumn{2}{c}{\textit{Average CPU(s)}} & & \textbf{3.44} & 879.93&320.3&326.01
    & & & \textbf{10.53} &879.08&	423.09&	511.27\\
    \cline{1-13}

  \end{tabular}
\end{table}

As revealed in Table~\ref{Comparaison} (and as already noticed in
Table~\ref{PerformanceCDADS}), on the whole, the total average of
\textit{\%dev} obtained by CDADS is 1.66\% and 6.39\% for the `Type-1'
and `Type-2' problems, respectively.  Compared to the corresponding
averages of 2.27\% and 7.28\% achieved by GA, and the corresponding
values of 5.39\% and 11.92 \% obtained by CP, CDADS outperforms the GA
and CP algorithms. Furthermore, CDADS was clearly superior to CP
especially for larger instances ($n=50$ and $n=100$).

As depicted in the table, MA finds slightly better solutions in
`Type-1' problems, that is 1.60\% is obtained by MA while CDADS gives
an average deviation percentage of 1.66\%. Overall, CDADS outperforms
significantly MA, as CDADS results are at 6.39\% from optimal
solutions (or lower bounds) for `Type-2' problems against 8.32\% for
MA. \\
 
To further assess the effectiveness of CDADS, we measure the number of
improved known solutions.
It can be seen from Table~\ref{Improvement} that CDADS improves $75$
known solutions among the $300$ tested instances. Thus, the rate of
improvement reaches $25$\%. The results also outline that most
improvements are spotted in large instances ($n=50,100$), see
figure~\ref{courbeAm}. No significant improvements are noticed for
small instances ($n=5,10$) since all optimal solutions for these
problems are known.

\begin{table}[ht]
  \centering
  \caption{Number of improved solutions}
  \label{Improvement}

  \begin{tabular}{c c c}
    \hline 
    $n$ &`Type-1' Problems & `Type-2' Problems \\
    \cline{1-3} \\[-2mm]
    $5$ & 0&0 \\
    \tabularnewline
    $10$ &  1&0 \\
   
    \tabularnewline
    $20$ &5&10 \\
   
    \tabularnewline
    $50$ &  8&20  \\
   
    \tabularnewline
    $100$ &8  &23  \\
   
    \cline{1-3}
   total&22&53 \\
    \cline{1-3}
 
  \end{tabular}
\end{table}
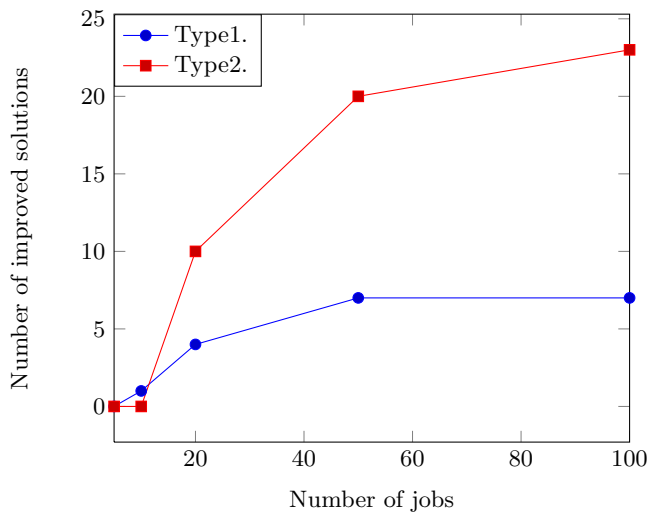
\begin{figure}[!h]
\centering

\begin{tikzpicture}
\begin{axis}[enlarge x limits=false,
xlabel={Number of jobs},
ylabel={Number of improved solutions},legend entries={Type1.,Type2.},legend style={at={(0,1)},anchor=north west}
]
\addplot coordinates {
(5,0) (10,1) (20,4)(50,7)(100,7)};
\addplot coordinates {
(5,0) (10,0) (20,10)(50,20)(100,23) };
\end{axis}
\end{tikzpicture}
\caption{Variation of the number of improved solutions with the number of jobs}
\label{courbeAm}
\end{figure}
In this study, we also compare the convergence of algorithms. It can
be seen from the last line of Table~\ref{Comparaison}, that CDADS
outperforms the genetic algorithm (GA), constraint programming (CP),
and the memetic algorithm (MA). Indeed, CDADS takes between less than
$0.1$ seconds (for small problems) and $47.3$ seconds (for large
problems) to find their solutions, while methods proposed in
\cite{jouglet09} converge much more slower [0.7 sec, 900 sec]. Even
all results were obtained under different computational budgets, we
can conclude that CDADS demonstrates fast convergence. Indeed,
according to Dongarra's normalized coefficients \cite{dongarra09}, our
machine is approximately only 3.5 times faster than the machine used
by Jouglet \emph{et al}.

\section{Conclusions}

In this paper, the hybrid flow shop problem with multiprocessor tasks
is addressed by means of a discrepancy search method. The proposed
method, Climbing Depth-bounded Adjacent Discrepancy Search (CDADS), is
based on adjacent discrepancies. We selected several heuristics to
generate the initial solution. A lower bound is also proposed to lead
a more efficient search. Compared to the best-known results in the
literature, CDADS provides better solutions in little CPU time.

In the short-term, we prospect to apply CDADS to simpler problems like
classical hybrid flow shop ($size_{ij}=1, \; \forall \, i,j$), widely
studied in the literature. Another expected aim would be to adapt the
proposed implementation of discrepancy search to more general
scheduling problems, in particular the Resource-Constrained Project
Scheduling Problem, which still remains one of the most challenging
problems in large-scale scheduling.

\bibliographystyle{plain}
\bibliography{biblio}

\end{document}